%% file: main.tex
\documentclass{llncs}
\usepackage{amsmath,graphicx}
\usepackage{amsmath,amssymb} 
\usepackage{epsfig}
\usepackage{multirow}
\usepackage{llncsdoc}
\usepackage{etoolbox}

\begin{document}

\mainmatter

\title{Efficient Active Learning for Image Classification and Segmentation using a Sample Selection and Conditional Generative Adversarial Network}

\titlerunning{Active Learning Using cGANs}

\author{Dwarikanath Mahapatra$^{1}$, Behzad Bozorgtabar$^{2}$, Jean-Philippe Thiran$^{2}$, Mauricio Reyes$^{3}$ }
 
\authorrunning{Mahapatra et. al.}
\institute{$^{1}$IBM Research Australia,\\
  $^{2}$Ecole Polytechnique Federale de Lausanne, Switzerland \\ 
	$^{3}$University of Bern, Switzerland \\ 
\email{dwarim@au1.ibm.com, behzad.bozorgtabar@epfl.ch, jean-philippe.thiran@epfl.ch, mauricio.reyes@istb.unibe.ch}}

 %



\maketitle

\begin{abstract}
Training robust deep learning (DL) systems for medical image classification or segmentation is challenging due to limited images covering different disease types and severity. We propose an active learning (AL)  framework to select most informative samples and add to the training data. We use conditional generative adversarial networks (cGANs) to generate realistic chest xray images with different disease characteristics by conditioning its generation on a real image sample. Informative samples to add to the training set are identified using a Bayesian neural network. Experiments show our proposed AL framework is able to achieve state of the art performance by using about $35\%$ of the full dataset, thus saving significant time and effort over conventional methods.
\end{abstract}

\input{AL_cGANs_MICCAI_Intro}

\input{AL_cGANs_MICCAI_Method}

\input{AL_cGANs_MICCAI_Expts}

\input{AL_cGANs_MICCAI_Concl}

\section*{Acknowledgement}

The authors acknowledge the support from SNSF project grant number $169607$.

\bibliographystyle{splncs03}
\bibliography{MICCAI2018_AL_Ref}

\end{document}

%% file: AL_cGANs_MICCAI_Intro.tex

\section{Introduction}
\label{sec:intro}

Medical image classification and segmentation are essential building blocks of computer aided diagnosis systems where deep learning (DL) approaches have led to state of the art performance \cite{DLRevTMI,Mahapatra_CVIU2019,Mahapatra_CMIG2019,Mahapatra_LME_PR2017,Zilly_CMIG_2016,Mahapatra_SSLAL_CD_CMPB,Mahapatra_SSLAL_Pro_JMI}. Robust DL approaches need large labeled datasets which is difficult for medical images because of: 1) limited expert availability; and 2) intensive manual effort required for curation.
 Active learning (AL) approaches overcome data scarcity with existing models by incrementally selecting the most informative unlabeled samples, querying their labels and adding them to the labeled set \cite{AL1,Mahapatra_LME_CVIU,LiTMI_2015,MahapatraJDI_Cardiac_FSL,Mahapatra_JSTSP2014,MahapatraTIP_RF2014,MahapatraTBME_Pro2014}. 
%
%
%
%
 AL in a DL framework poses the following challenges:1) labeled samples generated by current AL approaches are too few to train or finetune convolution neural networks (CNNs); 2) AL methods select informative samples using hand crafted features \cite{MahapatraMICCAI_CD2013,MahapatraTMI_CD2013,MahapatraJDICD2013,MahapatraJDIMutCont2013,MahapatraJDIGCSP2013,MahapatraJDIJSGR2013,MahapatraJDISkull2012}, while feature learning and model training are jointly optimized in CNNs. %

Recent approaches to using AL in a DL setting include Bayesian deep neural networks  \cite{Gal2017Active,MahapatraTIP2012,MahapatraTBME2011,MahapatraEURASIP2010,Kuanar_ICIP19,Bozorgtabar_ICCV19,Xing_MICCAI19}, leveraging separate unlabeled data with high classification uncertainty and high confidence for computer vision applications \cite{AL_CEAL,Mahapatra_ISBI19,MahapatraAL_MICCAI18,Mahapatra_MLMI18,Sedai_OMIA18,Sedai_MLMI18,MahapatraGAN_ISBI18}, and fully convolution networks (FCN)  for segmenting histopathology images \cite{YangAL_MICCAI17,Sedai_MICCAI17,Mahapatra_MICCAI17,Roy_ISBI17,Roy_DICTA16,Tennakoon_OMIA16,Sedai_OMIA16}.
%
%
%
We propose to generate synthetic data by training a conditional generative adversarial network (cGAN) that learns to generate realistic images by taking input masks of a specific anatomy. Our model is used with chest xray images to generate realistic images from input lung masks. This approach has the advantage of overcoming limitations of small training datasets by generating truly informative samples. We test the proposed AL approach for the key tasks of image classification and segmentation, demonstrating its ability to yield models with high accuracy while reducing the number of training samples.

%

%% file: AL_cGANs_MICCAI_Method.tex

\section{Methods}
\label{sec:met}

Our proposed AL approach identifies informative unlabeled samples to improve model performance. Most conventional AL approaches identify informative samples using uncertainty which could lead to bias as uncertainty values depend on the model. %
We propose a novel approach to generate diverse samples that can contribute meaningful information in training the model. Our framework has three components for: 1) sample generation; 2) classification/segmentation model; and 3) sample informativeness calculation.
%
 An initial small labeled set is used to finetune a pre-trained $VGG16$ \cite{VGG,Mahapatra_MLMI16,Sedai_EMBC16,Mahapatra_EMBC16,Mahapatra_MLMI15_Optic,Mahapatra_MLMI15_Prostate,Mahapatra_OMIA15,MahapatraISBI15_Optic,MahapatraISBI15_JSGR,MahapatraISBI15_CD,KuangAMM14,Mahapatra_ABD2014,Schuffler_ABD2014} (or any other classification/segmentation model) using  standard data augmentation (DA) through rotation and translation. The sample generator takes a test image and a manually segmented mask (and its variations) as input and generates realistic looking images (details in Sec~\ref{met:cGAN}). A Bayesian neural network (BNN) \cite{BDNN,MahapatraISBI_CD2014,MahapatraMICCAI_CD2013,Schuffler_ABD2013,MahapatraProISBI13,MahapatraRVISBI13,MahapatraWssISBI13,MahapatraCDFssISBI13} calculates generated images' informativeness and highly informative samples are added to the labeled image set.  
The new training images are used to fine-tune the previously trained classifier. The above steps are repeated till there is no change in classifier performance. %

\subsection{Conditional Generative Adversarial Networks}
\label{met:cGAN}

GANs \cite{GANs,MahapatraCDSPIE13,MahapatraABD12,MahapatraMLMI12,MahapatraSTACOM12,VosEMBC,MahapatraGRSPIE12,MahapatraMiccaiIAHBD11} learn a mapping from random noise vector \emph{z} to output image \emph{y}: $G : z \rightarrow y$. In contrast, conditional GANs (cGANs) \cite{CondGANS,MahapatraMiccai11,MahapatraMiccai10,MahapatraICIP10,MahapatraICDIP10a,MahapatraICDIP10b,MahapatraMiccai08,MahapatraISBI08} learn a mapping from observed image \emph{x} and random noise vector \emph{z}, to \emph{y}: $G :\{x,z\} \rightarrow y$. 
The generator $G$ is trained to produce outputs
that cannot be distinguished from``real'' images by an adversarially
trained discriminator, $D$. The cGAN objective function is:
\begin{equation}
L_{cGAN}=E_{x,y~p_{data}(x,y)} \left[\log D(x,y)\right] + E_{x~p_{data}(x),z~p_z(z)} \left[\log \left(1-D(x,G(x,z))\right)\right],
\label{eqn:cGan1}
\end{equation}
where $G$ tries to minimize this objective against $D$, that tries to maximize it, i.e. $G^{*}=\arg \min_G \max_D L_{cGAN}(G,D)$.
Previous approaches have used an additional $L2$ loss \cite{CGAN29,MahapatraICME08,MahapatraICBME08_Retrieve,MahapatraICBME08_Sal,MahapatraSPIE08,MahapatraICIT06,sZoom_Ar,CVIU_Ar} to encourage the generator output to be close to  ground truth in an $L2$ sense. We use $L1$ loss as it encourages less blurring \cite{MahapatraMICCAI_ISR,AMD_OCT,GANReg1_Ar,PGAN_Ar,Haze_Ar,Xr_Ar,RegGan_Ar,ISR_Ar}, and defined as:
\begin{equation}
L_{L1}(G)=E_{(x,y)~p_{data}(x,y),z~p_z(z)} \left[\left\|y-G(x,z)\right\|_1\right].
\label{eqn:cGan2}
\end{equation}
Thus the final objective function is :
\begin{equation}
G^{*}=\arg \min_G \max_D L_{cGAN}(G,D) + \lambda L_{L1}(G),
\label{eqn:cGan3}
\end{equation}
where $\lambda=10$, set empirically, balances the two components' contributions.

\subsubsection{Synthetic Image Generation:}
The parameters of $G$, $\theta_G$, are given by, %
\begin{equation}
\widehat{\theta}=\arg \min_{\theta_G} \frac{1}{N} \sum_{n=1}^{N} l\left(G_{\theta_G}(x,z),x,z\right),
\label{eq:theta1}
\end{equation}
$N$ is the number of images. Loss function $l$ combines content loss and adversarial loss (Eqn.~\ref{eqn:cGan1}), and $G(x,z)=y$. Content loss ($l_{content}$) encourages output image $y$ to have different appearance to $x$. $z$ is  the latent vector encoding (obtained from a pre-trained autoencoder) of the segmentation mask. $l_{content}$ is, 
\begin{equation}
l_{content} = NMI(x,y) - VGG(x,y) - MSE(x,y).
\label{eq:conLoss}
\end{equation} 
$NMI$ denotes the normalized mutual information (NMI) between $x$ and $y$, and is used to determine similarity of multimodal images. $VGG$ is the $L2$ distance between two images using all $512$ feature maps of Relu $4-1$ layer of a pre-trained $VGG16$ network \cite{VGG,LME_Ar,Misc,Health_p,Pat2,Pat3,Pat4,Pat5}. The VGG loss improves robustness by capturing information at different scales from multiple feature maps. $MSE$ is the intensity mean square error. For similar images, $NMI$ gives higher value while $VGG$ and $MSE$ give lower values. In practice $l_{content}$ is measuring the similarity (instead of dissimilarity in traditional loss functions) between two images, and takes higher values for similar images. Since we are minimizing the total loss function, $l_{content}$ encourages the  generated image  $y$ to be different from input $x$. %

The generator $G$ (Fig.~\ref{fig:Gan}(a)) employs residual blocks having two convolution layers with $3\times3$ filters and $64$ feature maps, followed by batch normalization and ReLU activation. It takes as input the test Xray image and the latent vector encoding of a mask (either original or altered) and outputs a realistic Xray image whose label class is the same as the original image. %
The discriminator $D$ (Figure~\ref{fig:Gan} (b)) has eight convolution layers with the kernels increasing by a factor of $2$ from $64$ to $512$ . Leaky ReLU is used and strided convolutions reduce the image dimension   when the number of features is doubled. The resulting $512$ feature maps are followed by two dense layers and a final sigmoid activation to obtain a probability map. $D$ evaluates similarity between $x$ and $y$. 
 To generate images with a wide variety of information we modify the segmentation masks of the test images by adopting one or more of the following steps:

\begin{figure}[t]
\begin{tabular}{cccc}
\includegraphics[height=2.9cm, width=6.99cm]{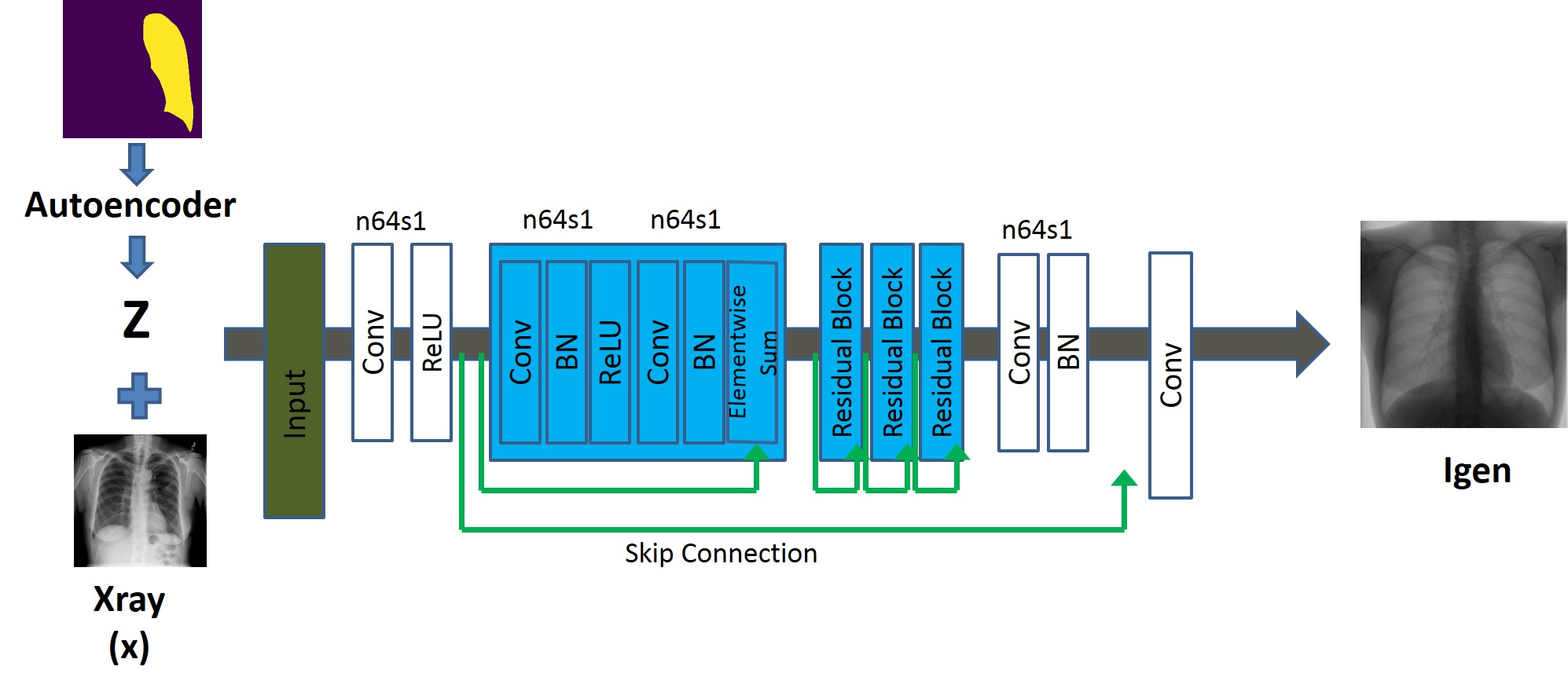}  & & &
\includegraphics[height=2.9cm, width=5.99cm]{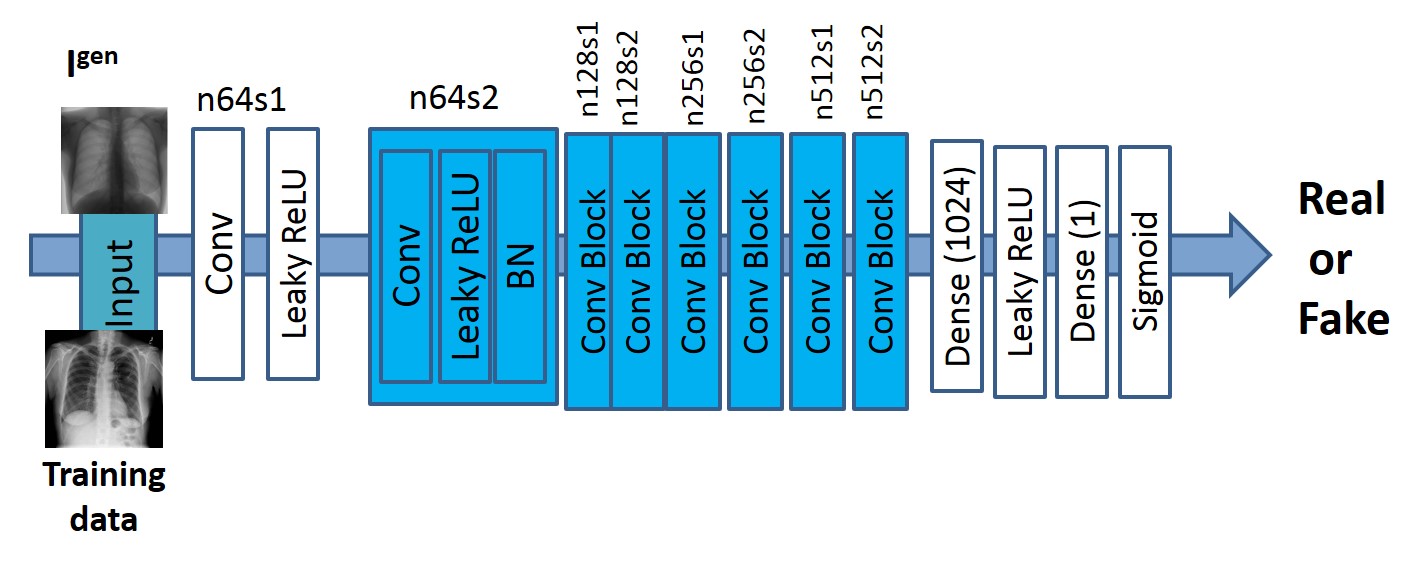}  \\
(a) & & & (b)\\
\end{tabular}
\caption{(a) Generator Network; (b) Discriminator network. $n64s1$ denotes $64$ feature maps (n) and stride (s) $1$ for each convolutional layer.}
\label{fig:Gan}
\end{figure}

\begin{enumerate}
\item \textbf{Boundary Displacement}: The boundary contours of the mask are displaced to change its shape. We select $25$ continuous points at multiple boundary locations, randomly displace each one of them by $\pm[1,15]$ pixels and fit a b-spline to change the boundary shape. The intensity of pixels outside the original mask are assigned by linear interpolation, or by generating intensity values from a distribution identical to that of the original mask. 

\item The intensity values of the lung region are changed by generating values from a uniform distribution modeled as $\alpha\mu + \beta\sigma$, where $\mu$ is the original distribution's mean, $\sigma$ is its  standard deviation, and $\alpha=[1,5],\beta=[2,10]$ (varied in steps of $0.2$).  %

\item Other conventional augmentation techniques like flipping, rotation and translation are also used.

\end{enumerate}

\begin{figure}[t]
\begin{tabular}{cccccc}
\includegraphics[height=1.8cm, width=1.8cm]{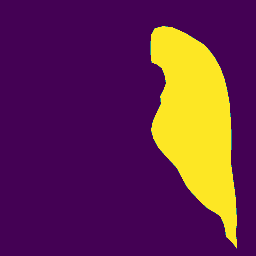}  &
\includegraphics[height=1.8cm, width=1.8cm]{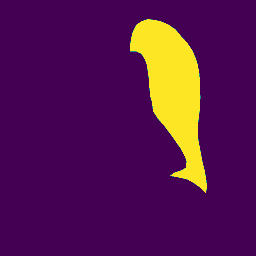}  &  
\includegraphics[height=1.8cm, width=1.8cm]{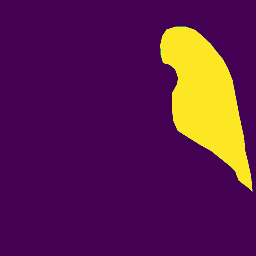}  &
\includegraphics[height=1.8cm, width=1.8cm]{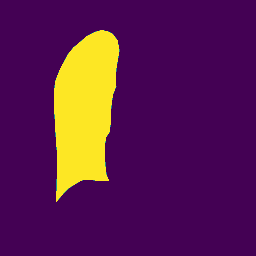}  &
\includegraphics[height=1.8cm, width=1.8cm]{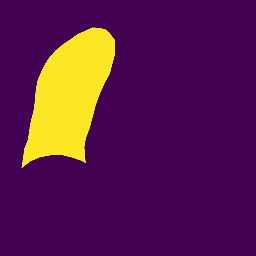}  &
\includegraphics[height=1.8cm, width=1.8cm]{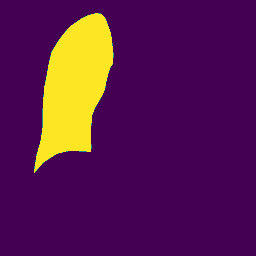}  \\
\includegraphics[height=1.8cm, width=1.8cm]{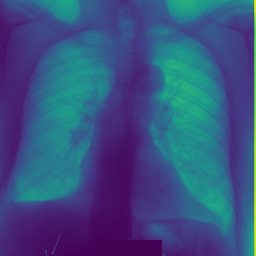}  &
\includegraphics[height=1.8cm, width=1.8cm]{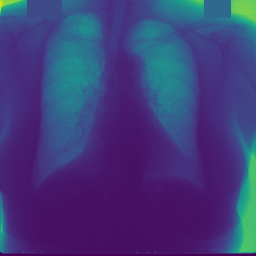}  &  
\includegraphics[height=1.8cm, width=1.8cm]{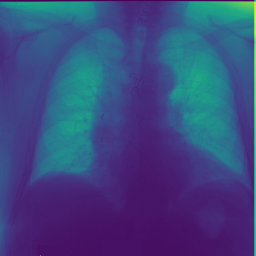}  &
\includegraphics[height=1.8cm, width=1.8cm]{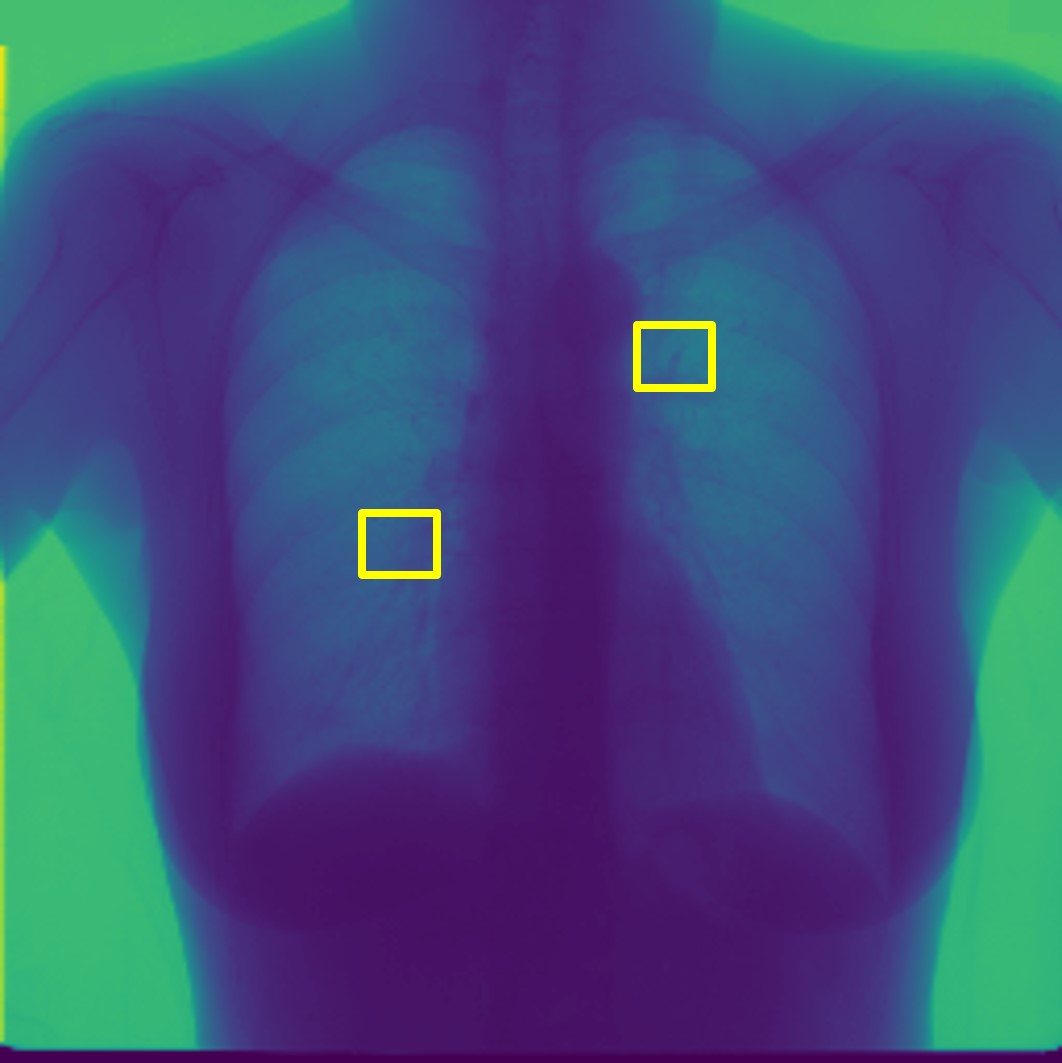}  &
\includegraphics[height=1.8cm, width=1.8cm]{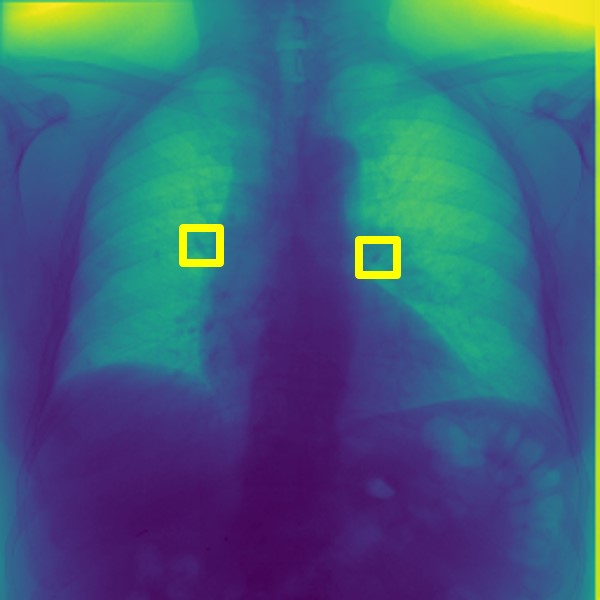}  &
\includegraphics[height=1.8cm, width=1.8cm]{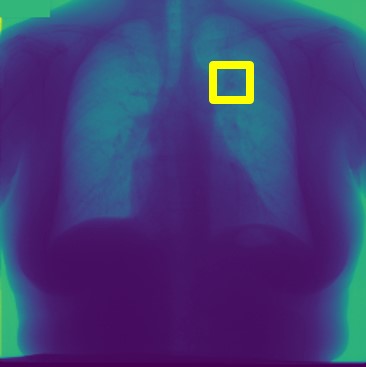}  \\
(a) & (b) & (c) & (d) & (e) & (f) \\
\end{tabular}
\caption{Mask (Top Row 1) and corresponding informative xray image (Bottom Row); (a)-(c) non-diseased cases; (d)-(f) images with nodules of different severity at the center of yellow box. (a), (d) are the original images while others are synthetic images generated by altering the mask characteristics.}
\label{fig:GenImages}
\end{figure}

For every test image we obtain up to $200$ synthetic images with their modified masks. 
Figure~\ref{fig:GenImages} (a) shows an original normal image (bottom row) and its mask (top row), and Figs.~\ref{fig:GenImages} (b,c) show generated `normal' images. Figure~\ref{fig:GenImages} (d) shows the corresponding image mask for an image with nodules, and Figs.~\ref{fig:GenImages} (e,f) show generated `nodule' images. Although the nodules are very difficult to observe with the naked eye, we highlight its position using yellow boxes. It is quite obvious that the generated images are realistic and suitable for training.

\subsection{Sample Informativeness Using Uncertainty form Bayesian Neural Networks}
\label{met:uncertainty}

Each generated image's uncertainty is calculated using the method described in \cite{BDNN,Pat6,Pat7,Pat8,Pat9,Pat10,Pat11}. Two types of uncertainty measures can be calculated from a Bayesian neural network (BNN). Aleotaric uncertainty models the noise in the observation while epistemic uncertainty models the uncertainty of model parameters.
 We adopt \cite{BDNN} to calculate uncertainty by combining the above two types. A brief description is given below and refer the reader to \cite{BDNN} for details.
For a BNN model $f$ mapping an input image $x$, to a unary output $\widehat{y}\in R$, the predictive uncertainty for pixel $y$ is approximated using:
\begin{equation}
Var(y)\approx \frac{1}{T} \sum_{t=1}^{T} \widehat{y}_t^{2} - \left(\frac{1}{T} \sum_{t=1}^{T} \widehat{y}_t \right)^{2} + \frac{1}{T} \sum_{t=1}^{T} \widehat{\sigma}_t^{2}
\label{eqn:Uncert}
\end{equation}
$\widehat{\sigma}^{2}_t$ is the BNN output for the predicted variance for pixel $y_t$, and ${\widehat{y}_t,\widehat{\sigma}^{2}_t}^{T}_{t=1}$ being a set of $T$ sampled outputs.

\subsection{Implementation Details}

Our initial network is a $VGG16$ network \cite{VGG} or $ResNet18$ \cite{ResNet} pre-trained on the Imagenet dataset. Our entire dataset had $93$ normal images and $154$ nodule images. We chose an initially labeled dataset of $16$ (chosen empirically) images from each class,  augment it $200$ times using standard data augmentation like rotation and translation, and use them to fine tune the last classification layer of the $VGG16$. The remaining test images and their masks were used to generate multiple images using our proposed cGAN approach ($200$ synthetic images for every test image as described earlier), and each generated image's uncertainty was calculated as described in Section~\ref{met:uncertainty}. 
We ranked the images with highest uncertainty score and the top $16$ images from each class were augmented $200$ times (rotation and translation) and used to further fine-tune the classifier. This ensures equal representation of normal and diseased samples in the samples to add to the training data. This sequence of steps is repeated till there is no further improvement of classifier accuracy when tested on a separate test set of $400$ images ($200$ images each of nodule and normal class). Our knowledge of image label allows quantitative analysis of model performance.

%% file: AL_cGANs_MICCAI_Expts.tex

\section{Experiments}
\label{sec:expts}

\paragraph{\textbf{Dataset Description}:}
 Our algorithm is trained on the SCR chest XRay database \cite{SCR} which has Xrays of $247$ ($93$ normal and $154$ nodule images, resized to $512\times512$ pixels) patients along with manual segmentations of the clavicles, lungs and heart. The dataset is augmented $500$ times using rotation, translation, scaling and flipping. We take a separate test set of $400$ images from the NIH dataset \cite{NIH} with $200$ normal images and $200$ images with nodules.

\subsection{Classification Results}

Here we show results for classifying different images using different amounts of labeled data and demonstrate our method's ability to optimize the amount of labeled data necessary to attain a given performance, as compared to conventional approaches where no sample selection is performed. In one set of experiments we used the entire training set of $247$ images and augmentation to fine tune the $VGG16$ classifier, and test it on the separate set of $400$ images. We call this the fully supervised learning (FSL) setting. Subsequently, in other experiments for AL we used different number of initial training samples in each update of the training data. The batch size is the same as the initial number of samples. 

 The results are summarized in Table~\ref{tab:AL1} where the classification performance in terms of sensitivity ($Sens$), specificity ($Spec$) and area under the curve ($AUC$) are reported for different settings using $VGG16$ and $ResNet18$ \cite{ResNet} classifiers. Under FSL, $5-$fold indicates normal $5$ fold cross validation; and $35\%$ indicates the scenario when $35\%$ of training data was randomly chosen to train the classifier and measure performance on test data (the average of $10$ such runs). We ensure that all samples were part of the training and test set atleast once.
In all cases AL classification performance reaches almost the same level as FSL when the number of training samples is approximately $35\%$ of the dataset. Subsequently increasing the number of samples does not lead to significant performance gain. This trend is observed for both classifiers, indicating it is not dependent upon classifier choice.

\begin{table*}[t]
\begin{tabular}{|c|c|c|c|c|c|c|c|c|c|c|c|c|c|c|}
\hline
{} & \multicolumn {10}{|c|}{Active learning ($\%$ labeled + Classifier)} & \multicolumn {4}{|c|}{FSL} \\  \hline
{} & \multicolumn {2}{|c|}{10\%}  & \multicolumn {2}{|c|}{15\%} & \multicolumn {2}{|c|}{25\%} & \multicolumn {2}{|c|}{30\%} & \multicolumn {2}{|c|}{35\%} & \multicolumn {2}{|c|} {$5$-fold} & \multicolumn {2}{|c|}{$35\%$} \\ \hline 
{} & {VGG16 \cite{VGG}} & {ResNet18 \cite{ResNet}}  & {\cite{VGG}} & {\cite{ResNet}} & {\cite{VGG}} & {\cite{ResNet}} & {\cite{VGG}} & {\cite{ResNet}}& {\cite{VGG}} & {\cite{ResNet}}& {\cite{VGG}} & {\cite{ResNet}} & {\cite{VGG}} & {\cite{ResNet}} \\ \hline 
{Sens} & {70.8} & {71.3}  & {75.3} & {76.2} & {89.2} & {89.7} & {91.5}  & {91.8} & {91.7} & {91.9} & {92.1} & {92.4} & {78.1} & {78.5}\\ \hline
{Spec} & {71.1} & {71.9}  & {76.0} & {76.8} & {89.9} & {90.5} & {92.1}  & {92.4} & {92.4} & {92.5} & {92.9} & {93.1} & {78.4} & {78.7} \\ \hline
{AUC} & {74.3} & {75.0}  & {78.7} & {79.4} & {92.5} & {93.0} & {94.9}  & {95.1} & {95.2} & {95.3} & {95.7} & {95.9} & {80.6} & {81.0} \\ \hline
{DM} & \multicolumn{2}{|c|}{68.2} & \multicolumn{2}{|c|}{74.1} & \multicolumn{2}{|c|}{86.4} & \multicolumn{2}{|c|}{90.4} & \multicolumn{2}{|c|}{91.0} & \multicolumn{2}{|c|}{91.3}  & \multicolumn{2}{|c|}{79.3} \\  \hline
{HD} & \multicolumn{2}{|c|}{18.7} & \multicolumn{2}{|c|}{14.3}  & \multicolumn{2}{|c|}{9.3} & \multicolumn{2}{|c|}{8.1} & \multicolumn{2}{|c|}{7.9} & \multicolumn{2}{|c|}{7.5} & \multicolumn{2}{|c|}{15.1} \\  \hline
\end{tabular}
\caption{Classification and Segmentation results for active learning framework of Xray images. DM-Dice metric and HD- Hausdorff distance}
\label{tab:AL1}
\end{table*}

\subsection{Segmentation Performance}

\begin{figure}[t]
\begin{tabular}{cccccc}
\includegraphics[height=1.8cm, width=1.8cm]{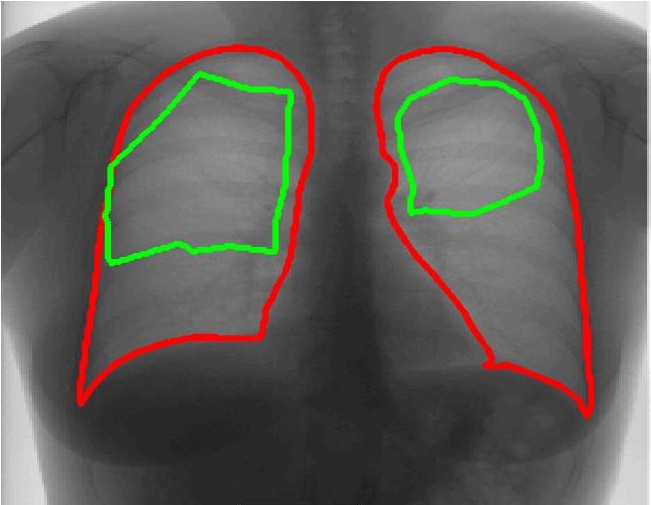}  &
\includegraphics[height=1.8cm, width=1.8cm]{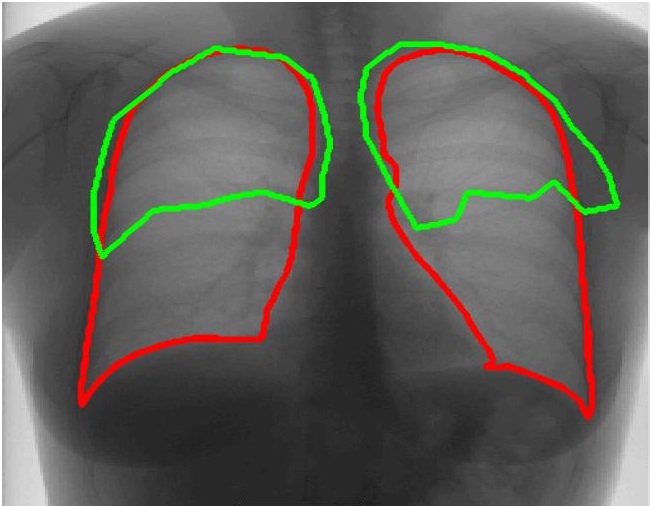}  &  
\includegraphics[height=1.8cm, width=1.8cm]{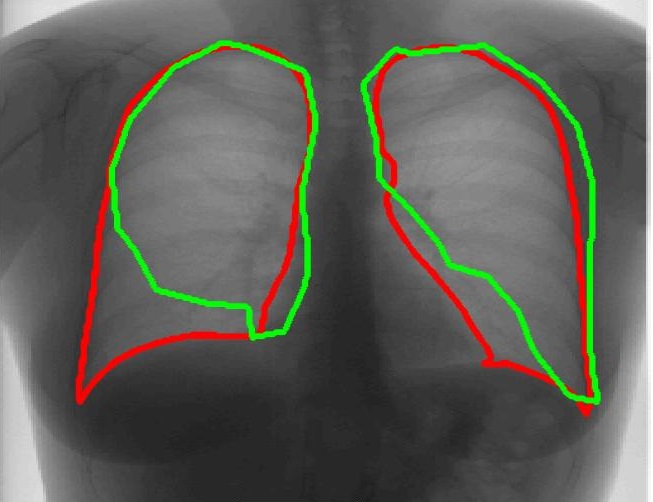}  &
\includegraphics[height=1.8cm, width=1.8cm]{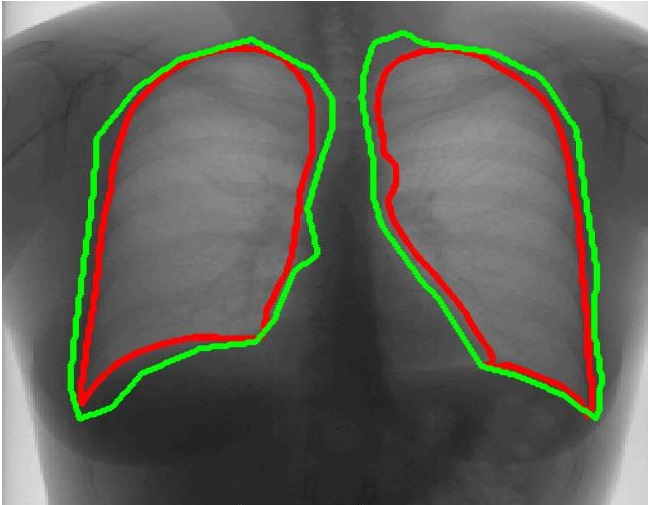}  &
\includegraphics[height=1.8cm, width=1.8cm]{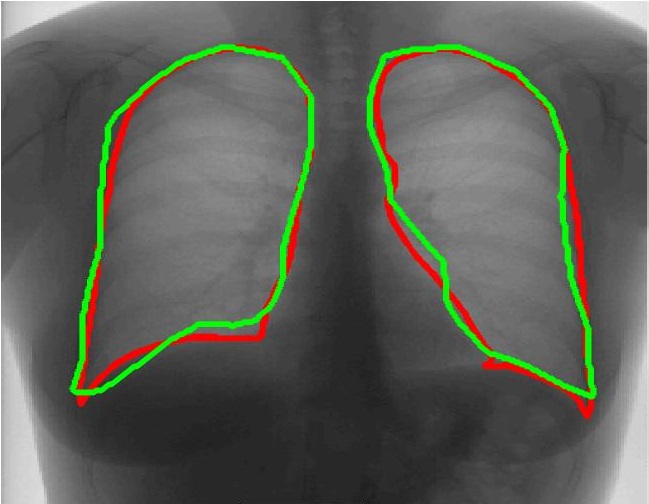}  &
\includegraphics[height=1.8cm, width=1.8cm]{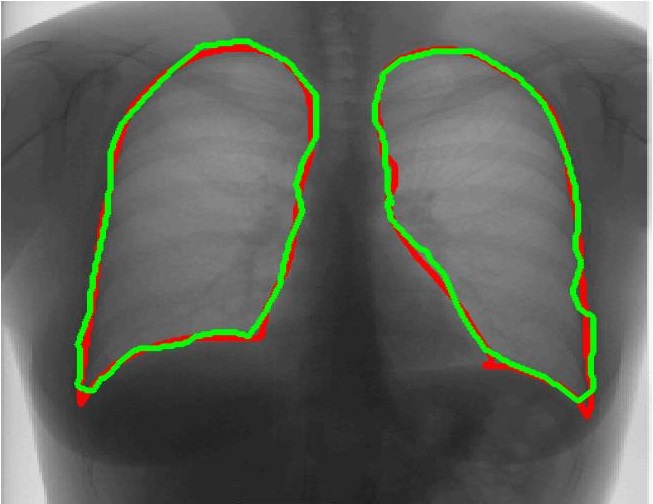}  \\
\includegraphics[height=1.8cm, width=1.8cm]{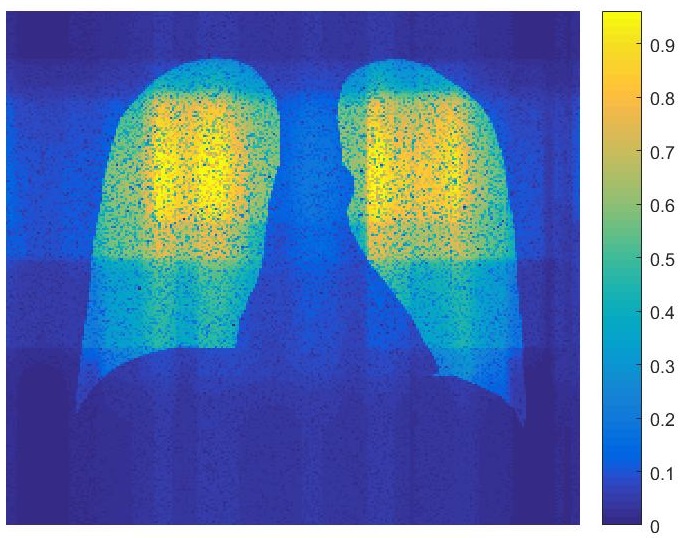}  &
\includegraphics[height=1.8cm, width=1.8cm]{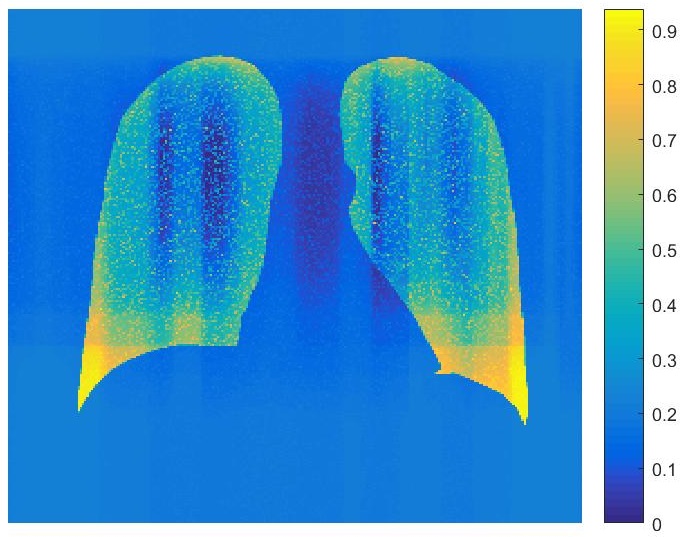}  &  
\includegraphics[height=1.8cm, width=1.8cm]{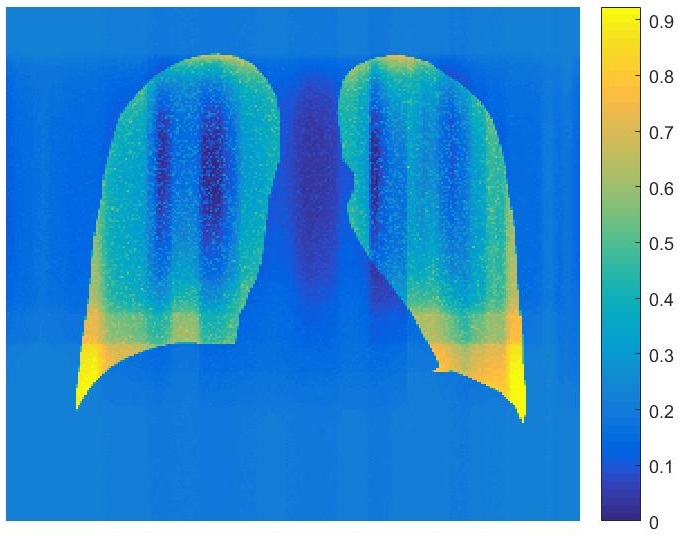}  &
\includegraphics[height=1.8cm, width=1.8cm]{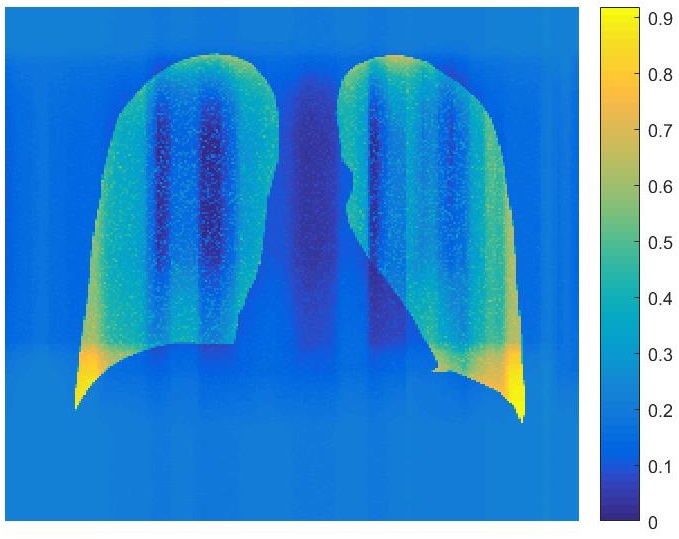}  &
\includegraphics[height=1.8cm, width=1.8cm]{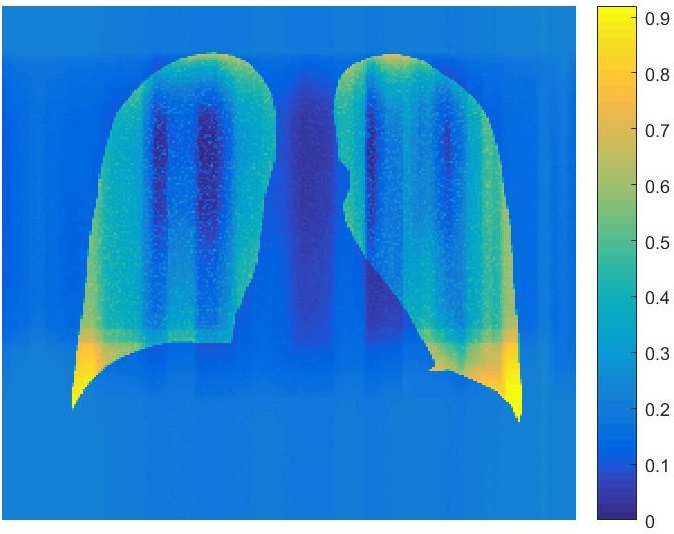}  &
\includegraphics[height=1.8cm, width=1.8cm]{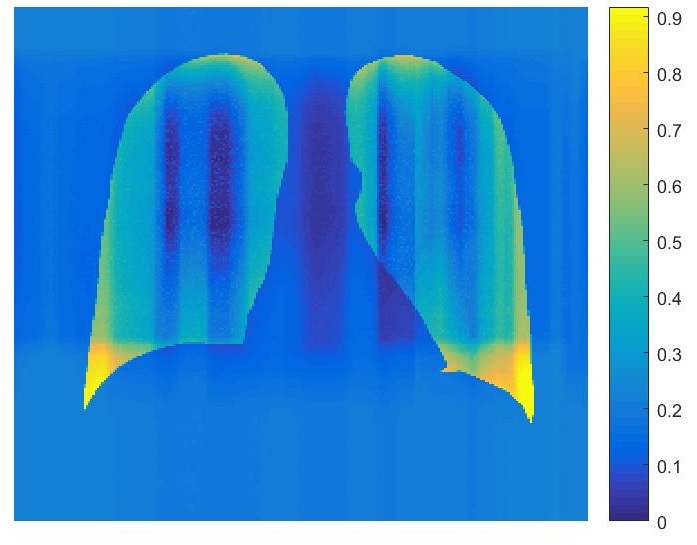}  \\
(a) & (b) & (c) & (d) & (e) & (f) \\
\end{tabular}
\caption{Segmentation (top row) and uncertainty map (bottom row) results for different numbers of labeled  examples in the training data (a) $5\%$; (b) $10\%$; (c) $20\%$; (d) $30\%$; (e) $35\%$; (f) $40\%$. Red contour is manual segmentation and green contour is the UNet generated segmentation.}
\label{fig:ALseg1}
\end{figure}

Using the labeled datasets at different stages we train a UNet \cite{Unet} for segmenting both lungs.  
The trained model is then evaluated on the separate set of $400$ images from the NIH database on which we manually segment both lungs. The segmentation performance for FSL and different AL settings is summarized in Table~\ref{tab:AL1} in terms of Dice Metric (DM) and Hausdorff Distance (HD). 
We observe that the segmentation performance reaches the same level as FSL at a fraction of the full dataset - in this case between $30-35\%$, which is similar for classification. Figure~\ref{fig:ALseg1} shows the segmentation results for different training models. When the number of training samples are less than $10\%$ the segmentation performance is quite bad in the most challenging cases. However the performance improves steadily till it stabilizes at the $35\%$ threshold.

\subsection{Savings in Annotation Effort}

Segmentation and classification results demonstrate that with the most informative samples, optimum performance can be achieved using a fraction of the dataset. This translates into significant savings in annotation cost as it reduces the number of images and pixels that need to be annotated  by an expert. We calculate the number of pixels in the images that were part of the AL based training set at different stages for both classification and segmentation. 
 At the point of optimum performance the number of annotated pixels in the training images is $33\%$. These numbers clearly suggest that using our AL framework can lead to savings of nearly $67\%$ in terms of time and effort put in by the experts.

%% file: AL_cGANs_MICCAI_Concl.tex

\section{Conclusion}
\label{sec:concl}

We have proposed a method to generate chest Xray images for active learning based model training by modifying the original masks of associated images.  %
A generated image's informativeness is calculated using a bayesian neural network, and the most informative samples are added to the training set. These sequence of steps are continued till there is no additional information provided by the labeled samples. Our experiments demonstrate that, with about $33-35\%$ labeled samples we can achieve almost equal classification and segmentation performance as obtained when using the full dataset. This is made possible by  selecting the most informative samples for training. Thus the model sees all the informative samples first, and achieves optimal performance in fewer iterations. The performance of the proposed AL based model translates into significant savings in annotation effort and clinicians' time. In future work we aim to  further investigate the realism of the generated images.